\documentclass[final]{l4dc2025}
\title[Controlling Participation in Federated Learning with Feedback]{Controlling Participation in Federated Learning with Feedback}
\usepackage{times}
\usepackage{algorithm}
\usepackage{algorithmic}
\usepackage{multirow}

\coltauthor{
  \Name{Michael Cummins}{\normalfont\textsuperscript{1}} \Email{m.cummins24@imperial.ac.uk}\\
  \Name{Guner Dilsad Er}{\normalfont\textsuperscript{2}} \Email{gder@tue.mpg.de}\\
  \Name{Michael Muehlebach}{\normalfont\textsuperscript{2}} \Email{michael.muehlebach@tue.mpg.de}\\
  \addr {\normalfont\textsuperscript{1}}Department of Electrical and Electronic Engineering, Imperial College London\\{\normalfont\textsuperscript{2}}Max Planck Institute for Intelligent Systems, Tübingen, Germany}

\usepackage{enumitem}  % Load the package for list customization
\usepackage{cleveref}
\usepackage{pgfplots}
\pgfplotsset{compat=1.17} % Specify the compatibility mode
\usepackage{tikz}
\usetikzlibrary{plotmarks}
\usetikzlibrary{arrows.meta,bending,positioning,calc} 
\usetikzlibrary[shapes,arrows]
\usepgfplotslibrary{fillbetween}

\pgfplotsset{
    width=4.5cm, 
    height=3.5cm,
    axis line style={black},
    tick style={black},
    grid style={dashed, gray},
    label style={font=\tiny},
    tick label style={font=\tiny},
    legend style={font=\tiny},
    title style={font=\tiny, align=center, yshift=-0.2cm},
    legend cell align={left},
    legend style={
        fill opacity=0.8,
        draw opacity=1,
        text opacity=1,
        draw=gray,
        fill=white,
        anchor=south east, % Anchor the legend at the southeast
        at={(0.97,0.03)}, % Position the legend at the bottom-right
        column sep=0.1ex, % Reduce the space between legend columns
        row sep=0.1ex, % Reduce the space between legend rows
        inner sep=0.1ex, % Reduce the inner padding of legend entries
        legend image post style={line width=0.3mm, shorten <=1.5mm, shorten >=1.5mm} % Shorten the legend lines
    },
}

\def\R{\mathbb{R}}

\def\prev{\text{prev}}

\def\S{\mathcal{S}}
\def\L{\mathcal{L}}
\def\I{\mathcal{I}}

\def\J{\mathrm{J}}

\newtheorem{asu}{Assumption}
\newtheorem*{lemma*}{Lemma}
\numberwithin{equation}{section}
\DeclareMathOperator*{\argmin}{argmin}

\newif\iflongversion
\newif\ifsubmit

\longversiontrue   %<<< Uncomment as required
\submitfalse  %<<< Uncomment as required

\begin{document}

\maketitle

\begin{abstract}%
We address the problem of client participation in federated learning, where traditional methods typically rely on a random selection of a small subset of clients for each training round. In contrast, we propose FedBack, a deterministic approach that leverages control-theoretic principles to manage client participation in ADMM-based cross-silo federated learning. FedBack models client participation as a discrete-time dynamical system and employs an integral feedback controller to adjust each client’s participation rate individually, based on the client's optimization dynamics. We provide global convergence guarantees for our approach by building on the recent federated learning research. Numerical experiments on federated image classification demonstrate that FedBack achieves up to 50\% improvement in communication and computational efficiency over algorithms that rely on a random selection of clients.\footnote{Open-source implementation of FedBack with code for running experiments is available at \url{https://github.com/michael-cummins/FedBack}}
\end{abstract}

\begin{keywords}%
 Federated Learning, Distributed Optimization, Client Participation Control, Event-Triggered Communication, ADMM, Feedback Control, Communication Efficiency
\end{keywords}

\section{Introduction}

The rapid growth of data generation across numerous devices has created new challenges for modern machine learning. Traditional centralized training methods, which involve aggregating data from individual devices into a central server for model training, are increasingly impractical due to both privacy concerns and the heavy communication costs associated with transferring large volumes of data. Federated learning (FL), a term coined by \citet{mcmahan_communicationefficient_2017}, provides a decentralized solution by allowing devices to collaboratively train machine learning models without the need to share raw data. Instead, each device trains a local model on its own data and transmits only updated model parameters to a central server or directly to other devices, thus addressing privacy concerns and facilitating large-scale learning across networks of distributed devices.

While FL effectively mitigates privacy risks, it introduces significant challenges in communication efficiency. The frequent exchange of model parameters between a central server and participating devices over potentially unreliable, bandwidth-constrained networks can be both energy-intensive and costly. This challenge is further amplified in large-scale networks, where high communication demands lead to increased operational costs and latency \citep{fed_challenges}. To address these issues, methods like FedAvg \citep{mcmahan_communicationefficient_2017} and FedProx \citep{fedprox}, and many more, have explored strategies such as random sampling of training clients to reduce communication overhead, offering partial solutions to this pressing concern. \looseness=-1

A promising approach to further reduce communication load in FL is event-triggered communication, where updates are transmitted only when significant changes occur in local model parameters \citep{fedevent}. Building on the sent-on-delta approach proposed by \citet{miskowicz_send_delta_2006}, we conceptualize client participation as an event-based control system \citep{astrom_event_2008, muehlebach2017distributed}, where clients participate only when necessary. Event-based methods, such as \citep{fedevent, zehtabi_decentralized_2022}, leverage peer-to-peer communication in decentralized settings with event-triggered mechanisms. These approaches achieve communication savings by transmitting model parameters only when certain conditions are met, such as changes in model parameters exceeding a fixed threshold. 
 Additionally, \cite{Chen_LAG_2018} adopt an event-triggered framework where lagged gradients are adaptively reused to reduce communication load. 
Building on these ideas, our strategy integrates event-triggered communication into an Alternating Direction Method of Multipliers (ADMM) framework, providing robustness to heterogeneity in local data distributions \citep{fedevent, zhou2023federated}.
Unlike existing event-based methods, which rely on static thresholds, our approach emphasizes the need for dynamically tuned thresholds that adapt to system state, network variability, and application-specific requirements, offering greater flexibility and efficiency.

Non-i.i.d. data distribution across clients creates a fundamental challenge in distributed learning by causing significant discrepancies among local datasets, which impairs the ability of individual models to generalize to a global solution. This divergence often results in slower convergence and suboptimal (global) models. Various approaches have been proposed to address this issue. Methods such as \citep{li_federated_2020, Acar_2021, Shi_2023} introduce proximal regularization terms to local objectives, which bias local updates toward the global model and reduce client drift. Similarly, ADMM-based methods, such as FedPD \citep{Zhang_2021} and FedADMM \citep{zhou2023federated, Wang_FedAdmm_2022, Gong_Li_Freris_2022}, align local models with global objectives through structured optimization formulations, though they differ in their strategies for client participation. Alternative splitting-based methods, such as the variant of Douglas-Rachford splitting proposed by \citet{feddr}, similarly align local and global objectives but remain constrained by random client participation. 
Meanwhile, FedNova \citep{wang_fednova_2020} addresses the inconsistency in server-side weighted aggregations, improving the alignment of local and global objectives. On another front, \citet{Karimireddy_SCAFFOLD_2020} introduce SCAFFOLD, which uses client control variates, serving as dual variables, to improve convergence in non-i.i.d. settings. However, SCAFFOLD comes with the trade-off of doubling communication costs, as it requires clients to exchange two variables, increasing the burden compared to other methods for the same participation rate.

In this paper, we introduce \textit{FedBack}, a novel FL framework that addresses these limitations through an adaptive thresholding mechanism designed to actively control device participation. Rather than relying on a fixed communication threshold, \textit{FedBack} dynamically adjusts the threshold $\delta$ based on factors such as network conditions, model accuracy requirements, and device capabilities. This tuning process allows \textit{FedBack} to balance the trade-off between communication cost and model convergence more effectively than fixed-threshold methods, achieving communication savings without compromising model performance. By incorporating an event-based communication strategy into an ADMM framework, \textit{FedBack} also enables joint optimization over both primal and dual variables, providing robustness against data and network heterogeneity.

Our approach is inspired by recent advances in the optimization literature that model algorithms as dynamical systems \citep{michael_2020, guanchun2023discrete, Nishihara_2015, dörfler2024systems,jordan_variational_2016, pavel_neurips}. The key idea is to interpret the sequence generated by an optimization algorithm as the trajectory of a dynamical system. For instance, \citet{dörfler2024systems} provides a systems-theoretic perspective on algorithms, illustrating examples such as interpreting primal-dual algorithms as proportional-integral controllers. 
% and gradient-based algorithms as closed-loop systems. 
Moreover, it  is  common to use similar mathematical tools, such as Lyapunov functions, for understanding and designing optimization algorithms with robust performance guarantees \citep{rawlings2017model, Nishihara_2015, Lessard_2016}.  Our contributions are as follows: 
\begin{enumerate}[label=\roman*), ref=\roman*, itemsep=0pt, topsep=0pt, parsep=0pt, partopsep=0pt,leftmargin=15pt]
\item We propose \textit{FedBack}, a novel cross-silo  FL algorithm that dynamically controls client participation by adjusting the communication threshold $\delta$ based on network conditions.
\item We establish global stability conditions for the feedback control law and provide global convergence guarantees for \textit{FedBack}.
\item Extensive numerical experiments are provided and illustrate \textit{FedBack} achieving substantial savings in both client communication and computation when compared to baseline methods, such as FedAvg \citep{mcmahan_communicationefficient_2017}, FedProx \citep{fedprox}, FedADMM \citep{zhou2023federated}, in communication efficiency and classification accuracy across various benchmark datasets. 
\end{enumerate}

\noindent
By providing a mechanism for adaptive participation control in FL, \textit{FedBack} addresses the core scaling challenges of large-scale, distributed learning (such as limited communication and computation resources) while ensuring robustness in non-i.i.d. settings. This makes \textit{FedBack} well-suited for deployment in real-world, bandwidth-constrained environments with varying data distributions.

The article is organized as follows: Sec.~\ref{sec:2} outlines the problem statement of learning from decentralized data via ADMM. A solution via feedback control is presented in Sec.~\ref{sec:3} and Sec.~\ref{sec:4} contains the theoretical analysis. Sec.~\ref{sec:5} provides a numerical study detailing the performance of \textit{FedBack} in two challenging benchmarks. We then conclude the paper in Sec.~\ref{sec:6} with final remarks and a comment on future work.

\section{Problem Statement}\label{sec:2}

We consider the optimization problem, 
\begin{equation}\label{eq:decentralised_ERM_condensed}
    \min_{\theta\in\R^d} \sum_{i=1}^N f_i(\theta),
\end{equation}
where $N$ is the number of clients in the distributed network and $f_i:\R^d \to \R$ is the loss function of client $i$ with dataset $D_i$. Each function $f_i$ has a Lipschitz continuous gradient with Lipschitz constant $r_i$.

We formulate an equivalent problem as 

\begin{align}\label{eq:ERM_admm_form}
    \min_{\{\theta_i\}_{1:N}, \omega} \: \sum_{i=1}^N f_i(\theta_i) \quad \text{s.t.}  \quad \theta_i = \omega \:\: \forall i \in \{1,\dots,N\},
\end{align}

\noindent
and solve \eqref{eq:ERM_admm_form} with ADMM \citep{boyd_distributed_2010} via the following dynamics, 

\begin{align}
    \lambda_i^{k+1} = &\lambda_i^k + \theta_i^{k} - \omega^{k}, 
    % \label{eq:GCADMM_dual}, 
    \quad \theta_i^{k+1} = \argmin_{\theta} f_i(\theta) + \frac{\rho}{2}|\theta - \omega^k + \lambda_i^{k+1}|^2, \label{eq:GCADMM_primal}\\
    \omega^{k+1} = & \frac{1}{N} \sum_{i=1}^N \left(\theta_i^{k+1} + \lambda_i^{k+1}\right), \label{eq:GCADMM_server}
\end{align} 

\noindent
where $\rho > 0$ is the proximal parameter,  $|\cdot|$ denotes the Euclidean distance and $\lambda_i^0 = 0$.  We note that the ADMM dynamics elicit a communication network where each client $i$ in the network downloads server parameters $\omega^k$ at round $k$, performs local computation to update local parameters $(\lambda_i^{k+1},\theta_i^{k+1})$ via  \eqref{eq:GCADMM_primal}\footnote{It is common (necessary for FedAvg \citep{mcmahan_communicationefficient_2017}) to warm start the optimization problem \eqref{eq:GCADMM_primal} with the server parameters $\omega^k$ received at that round. Although this is not a necessity in ADMM, it generally demonstrates superior empirical performance.}, and uploads $\lambda_i^{k+1} +\theta_i^{k+1}$ to the server to enable updates to the server parameters \eqref{eq:GCADMM_server}. However, it is well known that distributed optimization schemes such as ADMM do not scale well to cross-device or even cross-silo FL problems \citep{zhou2023federated, mcmahan_communicationefficient_2017, fedevent, silos}. We therefore follow a similar approach to \citep{fedevent} by only enforcing client participation when a certain event is triggered. As a result, we demonstrate dramatic improvement in communication and computation efficiency to vanilla ADMM while also maintaining global convergence guarantees.

\section{Federated Optimization with Controlled Participation}\label{sec:3}

A participation event in ADMM  takes place when the server sends the global parameters to a client $i$ and client $i$ sends its parameters to the server after doing some local computation, leading to $N$ communication events at each round $k$. This is usually guaranteed in base-splitting schemes for distributed optimization \citep{LSCO}, and motivates federated optimization to reduce the number of events necessary to reach a (local) minimizer of \eqref{eq:decentralised_ERM_condensed}. We address this issue by only selecting a client for participation if the difference between their local variables $(\lambda_i,\theta_i)$ and the server parameters $\omega$ exceeds a certain allowance $\delta\!\in\!\mathbb{R}_+$. Specifically, we define $z_i:=\theta_i+\lambda_i$ and the identifier function
\begin{align}\label{eq:identifier}
     S_i^k(\delta) := \begin{cases}
        1, &\text{if} \:\: |\omega^k - z_i^{\prev}| \geq \delta, \\
        0, &\text{else},
    \end{cases}
\end{align}
where $z_i^{\prev}$ is the most recent $z_i=\lambda_i+\theta_i$ received from client $i$. Thus, the identifier $S_i^k(\delta_i)=1$ if and only if client $i$ is selected to participate at round $k$. We note that with $\delta= 0$ we retrieve the original group consensus ADMM and $\delta \geq \delta_+$ (see Lemma~\ref{lem:bounded_delta} in Sec.~\ref{sec:4}) would create a deadlock in the network where no parameters would be updated. Our goal is therefore to find a sequence $\{\delta_i^k\}_{k=0}^{T-1}$ for each client $i$ such that
\begin{equation}\label{eq:true_load_dynamics}
    \lim_{T \to \infty} \frac{1}{T}\sum_{k=0}^{T-1} S_i^k(\delta_i^k) = \bar{L}_i,
\end{equation}
where $T$ is the total number of rounds necessary to reach a locally optimal solution of \eqref{eq:decentralised_ERM_condensed} and $\bar{L}_i$ is the desired participation rate of client $i$ over the course of the federated optimization procedure. We note that $\bar{L}_i$ may vary between clients as some may be willing to participate more/less frequently if they have access to more/less compute. In the forthcoming analysis, we assume $\bar{L}_i$ may differ between clients but only evaluate our method numerically for identical $\bar{L}_i$ in Sec.~\ref{sec:5}.

In the following, we propose a feedback control law that achieves \eqref{eq:true_load_dynamics}, where we consider $\delta_i$ to be the control input and $S_i^k(\delta_i^k)$ measurement output. Our feedback controller is as follows:
\begin{equation}\label{eq:integral_law}
    \delta_{i}^{k+1} = \delta_{i}^{k} + K(L^k_i - \bar{L}_i), 
\end{equation}
where $K\in\mathbb{R}_+$ is the control gain and $L_i^{k+1}$ is the output of a low pass filter applied to $S_i^k(\delta_i^k)$:
\begin{equation}\label{eq:load_dynamics}
    L_i^{k+1} = (1-\alpha)L_i^{k} + \alpha S_i^k(\delta_i^k),
\end{equation}
with time constant $\alpha\in (0,1)$ (typically ${\alpha \approx 0.9}$). The value of $L_i^k\in[0,1]$ therefore represents an estimate of how much client $i$ has communicated up to round $k$. Choosing a higher value for $\alpha$ will then emphasize recent measurements when computing $L_i^k$, whereas smaller $\alpha$ will emphasize the more distant past. 
We present the server-side event-based participation algorithm in Alg.~\ref{alg:select_participate},  whereby $\mathcal{S}_c$ denotes the client selection operator and $I_{\mathrm{s}}^k$ denotes the set of clients participating in round $k$. 

\begin{algorithm}[t]
\caption{Computing $\S_c$}
\label{alg:select_participate}
\begin{algorithmic}[0]
\STATE \textbf{Given:} $\mathcal{I} = \{1,\dots,N\}$, $\delta_i^k$, $\omega^k$, $z_i^{\prev}$
\STATE \textbf{Initialize:}  
$I_{\mathrm{s}}^k = \emptyset$, $\delta_i^0 = 0 \:\: \forall i \in \mathcal{I} $
\FOR{each client $i \in \I$}
    \IF{$S_i^k(\delta_i^k) = 1$}
        \STATE $I_{\mathrm{s}}^k \leftarrow I_{\mathrm{s}}^k \cup i$
    \ENDIF
    \STATE $L_i^{k+1} \leftarrow (1-\alpha)L_i^{k} + \alpha S_i^k(\delta_i^k)$ \hfill
    \COMMENT {Compute current running load for client "$i$"}
    \STATE $\delta_{i}^{k+1} \leftarrow \delta_{i}^{k} + K(L^k_i - \bar{L}_i)$  \hfill
    \COMMENT {Update threshold (control input)}
\ENDFOR
\RETURN $I_{\mathrm{s}}^k$
\end{algorithmic}
\end{algorithm}

Finally, we state the main result of the paper, the FedBack algorithm which utilizes Alg.~\ref{alg:select_participate} to reduce the number of participation events necessary to compute a locally optimal solution to \eqref{eq:decentralised_ERM_condensed}. 

\begin{algorithm}[H]
\caption{FedBack}
\label{alg:fedback}
\begin{algorithmic}[0]
    \STATE \textbf{Given:} $\rho$, clients $\mathcal{I} = \{1,\dots,N\}$, selection operator $\mathcal{S}_c$, accuracy $\{\varepsilon_k\}_{k=1}^T$
\STATE \textbf{Initialize:}  
\(\theta_1^0 = \dots = \theta_N^1 = z^0\), \(\lambda_i^0 = 0 \: \forall i \in \mathcal{I}\)
\FOR{each round $k = 0$ to $T-1$}
    \STATE Server selects $I_{\mathrm{s}}^k = \mathcal{S}_c(\mathcal{I})$
    \hfill \COMMENT{Compute via Alg.~\ref{alg:select_participate}}
    \STATE \textit{// Client-side computation}
    \FOR{each client $i \in I_{\mathrm{s}}^k$ in parallel}
    \STATE Download $\omega^k$ from server 
        \STATE \(\lambda_i^{k+1} \gets  \lambda_i^k + \theta_i^k - \omega^k\)\label{line:lambda_update}
        \STATE \(\theta_i^{k+1} \approx \argmin_{\theta \in \R^d} f_i(\theta) + \frac{\rho}{2} \left| \theta - \omega^k + \lambda_i^{k+1} \right|^2 \) \label{line:theta_update} \hfill \COMMENT{Local training: initialize  $\theta$ with $\omega^k$}
        \STATE \quad \quad \quad such that $| \nabla_\theta f_i(\theta^{k+1}_i) + {\rho} ( \theta^{k+1}_i - \omega^k + \lambda_i^{k+1})| \leq \varepsilon_k$
        \STATE \(z_i^{\text{prev}} \gets \lambda_i^{k+1} + \theta_i^{k+1}\)
        \STATE Upload $z_i^{\prev}$ to the server
    \ENDFOR
    \FOR{each client $i \notin I_{\mathrm{s}}^k$}
        \STATE  $(\theta_i^{k+1}, \lambda_i^{k+1}) \gets (\theta_i^k, \lambda_i^k)$ 
    \ENDFOR
    \STATE \textit{// Server-side computation}
    \STATE \(\omega^{k+1} \gets \frac{1}{N} \sum_{i=1}^{N} z_i^{\text{prev}}\) \label{line:global_update} \hfill
    \COMMENT{Update global average}
\ENDFOR
\end{algorithmic}
\end{algorithm}

\noindent
At each round $k$ of Alg.~\ref{alg:fedback}, the server selects a subset $I_{\mathrm{s}}^k$ of clients to participate via Alg.~\ref{alg:select_participate}. Each client $i$ in the set $I_{\mathrm{s}}^k$ performs the local  \eqref{eq:GCADMM_primal} update before uploading $z_i^{\prev}$ to the server.  The primal updates \eqref{eq:GCADMM_primal} can be performed inexactly, and we only require convergence to a stationary point with accuracy $\varepsilon_k$, where $\{\varepsilon_k\}$ is a positive sequence converging to zero. The server then aggregates each $z_i^{\prev}$ to compute $\omega^k$.
\looseness-1

Algorithm~\ref{alg:select_participate} requires that the server holds $z_i^{\prev}$ in memory for each client $i \in \I$. We acknowledge that this is not scalable in a cross-device setting \citep{mcmahan_communicationefficient_2017}. However, introducing dual variables already limits Alg.~\ref{alg:fedback} to the cross-silo setting \citep{silos} since each client has to maintain memory of $\lambda_i^k$, even when $i \notin I_{\mathrm{s}}^k$. Therefore, Alg.~\ref{alg:fedback} still remains scalable in the cross-silo setting ($\approx 100$ \textrm{clients}), provided that the models are not too large ($\approx$ 1 billion parameters). Future work may investigate compression strategies, such as the applications of the  Johnson-Lindenstrauss Lemma \citep{JL_lemma}, which could maintain scalability when dealing with even larger models.

We recall that $|\omega^k - z_i^{\prev}| = |x_i^{k} + \lambda_i^{k} - \omega^k|$,  which implies that the server selects clients with the largest proximal value in the primal update \eqref{eq:GCADMM_primal}.  Moreover, $|\lambda_i^{\prev} + \theta_i^{\prev} - \omega^k| = |\lambda_i^{k+1}|$ and $\lambda_i^{k+1} = \sum_{j=1}^{k} (\theta_i^j - \omega^j)$, implying that Alg.~\ref{alg:select_participate} naturally combats the client drift problem \citep{mcmahan_communicationefficient_2017,fedprox} by selecting clients that have built up a history of deviating too far from the server parameters. 

\section{Theoretical Analysis}\label{sec:4}
 
This section presents theoretical guarantees for Alg.~\ref{alg:fedback} under mild assumptions on the local objectives $f_i$.  We analyze the stability of participation dynamics and their limiting behavior.

\subsection{Global Stability of Participation Dynamics}

We first establish a key property of the participation dynamics, which ensures that the thresholds $\delta_i^k$  remain well-behaved over time. 
\vspace{-0.1cm}
\begin{lemma}[Bounded Threshold]\label{lem:bounded_delta}
Let the gradients in local training rounds \eqref{eq:GCADMM_primal} be bounded. Then, there exists a threshold value $\delta_+$, such that the identifier function $S_i^k$ in \eqref{eq:identifier} satisfies \begin{align*}
    S_i^k(\delta) = 0, \quad \forall \delta \geq \delta_+ > 0.
\end{align*}As a consequence, the following bound for the threshold at any time $k\geq 0$ holds, \begin{align*}
   \min\left\{\delta_i^0 - \frac{K}{\alpha},-K\left(\frac{1+\alpha}{\alpha}\right)\right\}\leq \delta_i^k\leq \max\left\{\delta_+ +K\left(\frac{1+\alpha}{\alpha}\right), \delta_i^0+\frac{K}{\alpha}\right\}.
\end{align*}
\end{lemma}
\vspace{-0.1cm}
\begin{proof}
 See \iflongversion Appendix \ref{proof:bounded} for the proof. \fi \ifsubmit \citep{Cummins_2024_long_version} for the proof. \fi 
\end{proof}
\noindent
We now show that the participation rates converge to the target value $\bar{L}_i$ under the closed-loop dynamics.

\begin{theorem}[Global Stability]\label{thm:stable}
    Let the sequence $\{\delta_i^k\}_{k=0}^{T-1}$ be generated by the closed loop dynamics \eqref{eq:integral_law} and \eqref{eq:load_dynamics}. Then, the time-averaged participation rate $\frac{1}{T}\sum_{k=0}^{T-1} S_i^k(\delta_i^k)$ converges to the target value $\bar{L}_i$ at a rate of $\mathcal{O}\left(\frac{1}{T}\right)$, for any $\bar{L}_i \in [0,1]$ and for $K>0$.
\end{theorem}

\begin{proof}
    By rearranging \eqref{eq:load_dynamics}, we express the dynamics of $L^k_i$ as $ L_i^{k+1} - L_i^k = -\alpha L_i^k + \alpha S_i^k(\delta_i^k) $, which concludes  %% 
    \vspace{-0.2cm}\begin{equation}\label{eq:telescopic}
   L_i^{T} - L_i^0 = -\alpha \sum_{k=0}^{T-1} L_i^k + \alpha \sum_{k=0}^{T-1} S_i^k(\delta_i^k). 
    \end{equation}%%
    In a similar fashion, we rearrange \eqref{eq:integral_law} 
    \begin{equation}\label{eq:sum_Lik}
       \sum_{k=0}^{T-1}L_i^k = \frac{\delta_i^T - \delta_i^0}{K} + T\bar{L}_i.
    \end{equation}
    Combining \eqref{eq:telescopic} and \eqref{eq:sum_Lik},  we have 
    \begin{align*}%\label{eq:second_last}
        \frac{1}{T} \sum_{k=0}^{T-1} S_i^k(\delta_i^k) = \bar{L}_i + \frac{\delta_i^T - \delta_i^0}{KT} + \frac{L_i^{T} - L_i^0}{\alpha T}. 
    \end{align*}
   Using boundedness of $\delta_i^T$ in Lemma~\ref{lem:bounded_delta} and the fact $|L_i^T - L_i^0| \leq 1$, we establish upper and lower bounds as
\begin{align*}
    \frac{c_1}{T} \leq   \frac{1}{T} \sum_{k=0}^{T-1} S_i^k(\delta_i^k) - \bar{L}_i \leq \frac{c_2}{T},
\end{align*}with constants $c_1= \min\left\{-\frac{2}{\alpha },-\frac{\delta^0_i}{K}-\frac{(2+\alpha)}{\alpha }\right\}$ and $c_2=\max\left\{\frac{\delta_+-\delta^0_i}{K}+\frac{(2+\alpha)}{\alpha },\frac{(2+\alpha)}{\alpha }\right\}$, which yields the desired result.
\end{proof}
\vspace{-0.5cm}

\begin{remark}
    Although we explicitly devote our attention to ADMM, we note that global stability is independent of the deployed decomposition algorithm. The distance metric utilized in \eqref{eq:identifier} can be designed freely as long as client gradients are bounded in local training rounds. 
\end{remark}

\vspace{-0.5cm}
\subsection{Global Convergence}

To analyze global convergence, we define the global Lagrangian $\L$ and corresponding local Lagrangians $\L_i$ of \eqref{eq:ERM_admm_form} as
\begin{equation}\label{eq:lagrangian}
    \L(\omega,\Theta,\Lambda) := F(\Theta) + \sum_{i=1}^N \left( \lambda_i^\intercal(\theta_i - \omega) + \frac{\rho}{2}|\theta_i - \omega|^2 \right) = \sum_{i=1}^N \L_i(\omega, \theta_i, \lambda_i),
\end{equation}
\noindent
where $\Theta := (\theta_1,\dots,\theta_N)$, $\Lambda := (\lambda_1,\dots,\lambda_N)$ and ${F(\Theta) := \sum_{i=1}^N f_i(\theta_i)}$. By Lemma~\ref{lem:communicate} (below), we show that none of the clients will stop participating in the optimization. This result guarantees global convergence to a stationary point of \eqref{eq:lagrangian} by applying the result of \citep{zhou2023federated}.

\begin{lemma}[Communication Guarantee]\label{lem:communicate}
    Let $K>0$ and $\bar{L}_i >0$, then $\underset{k \to \infty}{\lim \sup}$ $ \:S_i^k(\delta_i^k) = 1$.
\end{lemma}
\vspace{-0.2cm}
\begin{proof}
 See \iflongversion Appendix \ref{proof:communicate} for the proof. \fi \ifsubmit     \citep{Cummins_2024_long_version} for the proof. \fi 
\end{proof}
\vspace{-0.2cm}

\noindent
Lemma~\ref{lem:communicate} shows that FedBack satisfies one of the primary requirements for global convergence. We now state the assumptions on \eqref{eq:lagrangian}. 

\begin{asu}\label{as:coercive}
    The objective function $F(\Theta)$ is coercive, i.e., $F(\Theta)\rightarrow \infty$ as $|\Theta|\rightarrow \infty$.
\end{asu}
\begin{asu}\label{as:penalty}
    The parameter $\rho\!\in\!\R^+$ satisfies $\rho\!\geq\!\max\left\{\frac{3n_1r_1}{n},\dots,\frac{3n_Nr_N}{n}\!\right\}$, where $n$ is the number of data points in the global dataset and $n_i$ is the number of data points in the $\text{i}^\text{th}$ local dataset.
\end{asu}

\begin{theorem}[Global Convergence]
    Let Assumptions \ref{as:coercive} and \ref{as:penalty} hold and let $K>0$ and $\bar{L}_i\in (0,1]$. 
    The following statements hold
    \begin{enumerate}[itemsep=0pt, topsep=0pt, parsep=0pt, partopsep=0pt,leftmargin=15pt]
        \item The sequence $\{(\omega^k,\Theta^k,\Lambda^k)\}$ is bounded.
        \item The sequences $\{\L(\omega^k,\Theta^k,\Lambda^k)\}$, $\{F(\Theta^k)\}$ and $\{f(\omega^k)\}$ all converge to the same value:%%
        \begin{align*}
            \lim_{k \to \infty} \L(\omega^k,\Theta^k,\Lambda^k) = \lim_{k \to \infty} F(\Theta^k) =  \lim_{k \to \infty} \sum_{i=1}^N f_i(\omega^k).
        \end{align*}%%
        \item $\nabla\!F(\Theta^k)$ and $\nabla\!f_i(\omega^k)$ eventually vanish, i.e., $\lim_{k \to \infty}\!\nabla\! F(\Theta^k)\!=\!\lim_{k \to \infty}\!\sum_{i=1}^N \!\nabla \!f_i(\omega^k)\!=\!0$.
        \item Any accumulating point $(\omega^\infty, \Theta^\infty, \Lambda^\infty)$ of $\{(\omega^k,\Theta^k,\Lambda^k)\}$ is a stationary point of \eqref{eq:lagrangian}.
    \end{enumerate}    
\end{theorem}
\begin{proof}
    As a result of Lemma~\ref{lem:communicate}, the result of \citep{zhou2023federated} holds. This yields the desired conclusion.
\end{proof}
\vspace{-0.2cm}

\noindent
 We refer the reader to \cite{fedevent} for results on the convergence rate of ADMM with event-based communication. Due to the upper bound on $\delta_i^k$, the results from \cite{fedevent} apply directly. 
\vspace{-0.7cm}
\section{Numerical Evaluation}\label{sec:5}

We evaluate FedBack against FedAvg \citep{mcmahan_communicationefficient_2017}, FedProx \citep{fedprox}, and FedADMM \citep{zhou2023federated} on MNIST and CIFAR-10 classification tasks, using non-i.i.d. data distributions across 100 clients. FedADMM is a natural comparison for FedBack and a version of FedAvg/FedProx may be recovered from FedADMM by enforcing $\rho=0$ and $\lambda_i^{k+1}=0$ respectively and performing a non-weighted aggregation on the server-side. 

\begin{figure}[h!]
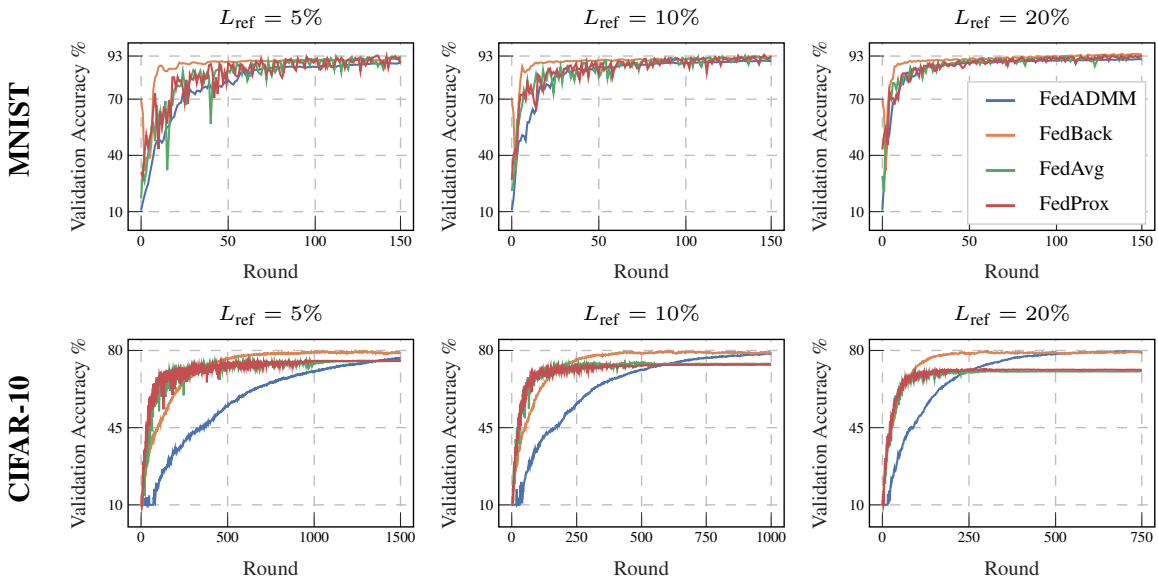

\centering
    % First row with "MNIST"
    \begin{minipage}[t]{0.03\linewidth} % Narrow minipage for the vertical text
  
    \rotatebox{90}{\hspace{2cm}\textbf{MNIST}} % Rotate text vertically
    
    \end{minipage}
    \hfill
    \begin{minipage}[b]{0.31\linewidth} % Adjust width to 0.32 for tighter fit
        \input{tex_figs/mnist/rate_5}
    \end{minipage}
    \hfill
    \begin{minipage}[b]{0.31\linewidth}
        \input{tex_figs/mnist/rate_10}
    \end{minipage}
    \hfill
    \begin{minipage}[b]{0.31\linewidth}
        \input{tex_figs/mnist/rate_20}        
    \end{minipage}     
    \hfill
      % Second row with "CIFAR-10"
    \begin{minipage}[t]{0.03\linewidth} % Narrow minipage for the vertical text
   
    \rotatebox{90}{\hspace{1.6cm}\textbf{CIFAR-10}} % Rotate text vertically
    
    \end{minipage}
    \hfill
    \begin{minipage}[b]{0.31\linewidth} % Adjust width to 0.32 for tighter fit
     \vspace{-0.5cm}
        \input{tex_figs/cifar/rate_5}
    \end{minipage}
    \hfill
    \begin{minipage}[b]{0.31\linewidth}
     \vspace{-0.5cm}
        \input{tex_figs/cifar/rate_10}
    \end{minipage}
    \hfill
    \begin{minipage}[b]{0.31\linewidth}
     \vspace{-0.5cm}
        \input{tex_figs/cifar/rate_20}
    \end{minipage}
    
    \vspace{-1cm}
    \caption{Validation accuracy of server parameters $\omega^k$ per round $k$ for MNIST (top row) and CIFAR-10 (bottom row) classifiers for each FL algorithm with communication load references $\bar{L}\!=\!\{0.05,0.1,0.2\}$. For FedADMM, FedAvg and FedProx, we randomly sample an $\bar{L}$ proportion of clients, uniformly at random, for participation at each round. }
    \label{fig:val_acc}
\end{figure} 

\textbf{Metrics}~ Due to the nature of Alg.~\ref{alg:select_participate}, there will be a varying number of client's participating in each round of Alg.~\ref{alg:fedback}. We therefore consider the total number of participation events necessary to achieve a desired accuracy as our first metric.  When evaluating the performance of FedADMM, FedAvg and FedProx, this metric is equivalent to the number of rounds required to reach a specified accuracy. Furthermore, we track the validation accuracy per round for each algorithm for a select $\bar{L} = \bar{L}_i \: (\forall i \in \I)$ to illustrate the superior convergence of FedBack and the advantage of using a deterministic client selection scheme. Finally, we support Thm.~\ref{thm:stable} by analyzing the realized participation rate among clients for a given $\bar{L}$.

\textbf{MNIST}~ Client data is distributed such that each client has an equal number of data points but is restricted to two unique digits. The classifier is a fully connected neural network with a single hidden layer of 200 neurons and ReLU activation.  The expected accuracy of the classifier is $93\%$ when training in a centralized fashion. Local updates are performed using SGD with learning rate 0.01, momentum factor 0.9 and batch size 42 for two epochs. We choose $K=2$ and $\alpha=0.9$.

\textbf{CIFAR-10}~ Client data is distributed using a Dirichlet distribution \citep{moon, bayesianfl, matched_average} with concentration parameter $\beta=0.5$. The classifier is a CNN with three convolutional layers, three fully connected layers and ReLU activation. When training in the centralized case, the expected test accuracy is $80\%$. Local updates are performed using SGD with learning rate 0.01, momentum factor 0.9 and batch size 20 for four epochs. We again choose $\alpha=0.9$ but increase $K=5$, since the CIFAR-10 classifier contains significantly more parameters.

\noindent
All experiments are implemented in PyTorch with Nvidia A100 GPUs. Fig.~\ref{fig:val_acc} demonstrates a notable trade-off regarding the noise in the server parameters, particularly for FedProx and FedAvg. This noise can be attributed to the random sampling of clients when $\bar{L}$ is low, creating high variance in the server parameters between rounds. This is particularly unfavorable in practical FL scenarios where the server does not have access to a validation dataset. To ensure a low level of variance in $\omega^k$ between rounds, one would have to enforce a high $\bar{L}$ or run the FL procedure for longer than required to ensure convergence. In contrast, FedBack uses an adaptive participation mechanism, which reduces the variance in global model parameters, leading to more stable performance over time when dealing with a low $\bar{L}$.

\begin{table}[t]
    \centering
    \begin{tabular}{|c|c|c|c|c|c|c|c|c|}
    \hline
       & Algorithm  & $\bar{L}\!=\! 5\%$& $\bar{L}\!=\! 10\%$& $\bar{L} \!=\! 15\%$& $\bar{L}\!=\! 20\%$ & $\bar{L}\!=\! 40\%$ & $\bar{L} \!=\!60\%$\\
        \hline
        \multirow{4}{*}{\rotatebox{90}{{MNIST}}} & \textbf{FedBack} & 412 & \textbf{430} & \textbf{493}& \textbf{538} & \textbf{677} &  1274\\
      &  \textbf{FedADMM} & $>750$ & 1280& 1215& 1340 & 1160 & \textbf{1080} \\
       & \textbf{FedAvg} & 370 & 550& 675& 720 & 1280 & 2160\\
        
       & \textbf{FedProx} & \textbf{335} & 620& 780& 880 & 1240 & 2400 \\
        \hline
         \multirow{4}{*}{\rotatebox{90}{{CIFAR10}}}    &\textbf{FedBack} & \textbf{3936} & \textbf{4167} & \textbf{4328}& \textbf{4402}& \textbf{4659}&\textbf{5064}\\       
     &   \textbf{FedADMM} & $>7500$  & 8960& 9720& 9260 & 9000 & 9840\\
      &  \textbf{FedAvg} & N/A & N/A& N/A& N/A & N/A & N/A\\
      &  \textbf{FedProx} & N/A & N/A& N/A& N/A & N/A & N/A\\\hline  
    \end{tabular}
    \vspace{-0.1cm}
    \caption[MNIST Accuracy]{Total number of participation events necessary for each algorithm to reach the target accuracy of 90\% for the MNIST Classifier and 78\% for the CIFAR-10 Classifier. The “N/A” entries represent cases where the algorithm could not achieve the desired accuracy within the given number of rounds. }
    \label{tab:num_events} 
\end{table}

We see from Tab.~\ref{tab:num_events} that FedBack outperforms all algorithms in the majority of test cases regarding participation efficiency across both MNIST and CIFAR-10 datasets. In the case of MNIST, FedProx is the most efficient at $\bar{L}=5\%$ but FedBack remains close with only 412 participation events and significantly less variance, as illustrated in Fig.~\ref{fig:val_acc}. CIFAR-10 classifiers trained with FedAvg and FedProx fail to reach the desired accuracy. This is likely due to being a heuristic modification of the distributed gradient method \citep{mcmahan_communicationefficient_2017,LSCO}, which is used to solve \eqref{eq:ERM_admm_form}. 

\begin{table}[H]
    \centering
    \begin{tabular}{|c|c|c|c|c|c|c|c|c|}
    \hline
       & $\bar{L}\!=\! 5\%$& $\bar{L}\!=\! 10\%$& $\bar{L} \!=\! 15\%$& $\bar{L}\!=\! 20\%$ & $\bar{L}\!=\! 40\%$ & $\bar{L} \!=\!60\%$\\
       \hline
        \textbf{CIFAR-10} & 5.54\% & 10.67\% & 15.81\% & 20.62\% & 40.71\% & 60.49\% \\
        \hline
        \textbf{MNIST} & 7.46\% & 12.07\% & 17.41\% & 22.58\% & 43.03\% & 64.93\% \\
        \hline
    \end{tabular}
    \vspace{-0.3cm}
    \caption{Average participation rate of clients in the network for a given $\bar{L}$. We compute the percentage of how many rounds each client participates in across the entire FL procedure and compute the average.}
    \label{tab:final_load}
\end{table}

\noindent
In addition, our feedback policy proved particularly effective in stabilizing the communication load throughout the learning process. Tab.~\ref{tab:final_load} demonstrates the communication load stabilizing within 0.9\% precision over longer training rounds in the CIFAR-10 experiments. This is a significant improvement over the MNIST experiments, where the dynamics did not have sufficient time to stabilize, highlighting FedBack's potential for long-term, stable communication management in federated learning settings. Although our feedback policy demonstrated less accurate tracking of $\bar{L}$, Tab.~\ref{tab:num_events} still demonstrates superiority in reaching the desired accuracy in much fewer participation events for $\bar{L}\!=\! \{0.10,0.15,0.20,0.40\}$.

\section{Conclusion}\label{sec:6}
This paper introduced FedBack, a modification of the cross-silo  FedADMM algorithm that applies a control theoretic methodology to client participation. We presented a theoretical analysis that justified our approach while also maintaining previously established global convergence properties. FedBack demonstrated a great advantage over FedAvg, FedProx and FedADMM during numerical evaluation. In most cases, FedBack was able to achieve the desired accuracy in almost half the number of participation events, while also demonstrating a drastic reduction in the variance of server parameters for a low $\bar{L}$. Moreover, our feedback control law was evaluated by accurately tracking $\bar{L}$ to an exceptional degree in the CIFAR-10 experiments and an acceptable degree in the MNIST experiments. Both FedBack and FedADMM were able to achieve the same validation accuracy as a model trained in a fully centralized fashion, with FedBack demonstrating a much faster convergence rate in the CIFAR-10 experiments.

In conclusion, by dynamically adjusting the communication threshold $\delta_i$ for each client, FedBack achieves better control over the communication load while maintaining stability in model training. This is particularly beneficial in practical FL scenarios where the communication resources are limited and/or changing over time. Further work will investigate more advanced feedback strategies and possibly integrate the feedback mechanism to other FL algorithms which utilize proximal terms in their objective, such as FedProx.

\acks{The authors thank Harun Siljak and Anthony Quinn for engaging in helpful  discussions, and  particularly thank Harun for enabling the collaboration. Guner Dilsad Er and Michael Muehlebach thank the German Research Foundation and the International Max Planck Research School for Intelligent Systems (IMPRS-IS) for their support.}

\bibliography{main}
\iflongversion 
\newpage

\appendix
\section{Supporting Lemmas}

\subsection{Proof of Lemma~\ref{lem:bounded_delta}}\label{proof:bounded}
We restate Lemma~\ref{lem:bounded_delta} for the convenience of the reader.
\begin{lemma*}
Let the gradients in local training rounds \eqref{eq:GCADMM_primal} be bounded. Then, there exists a threshold value $\delta_+$, such that the identifier function $S_i^k:\R\rightarrow\{0,1\}$ in \eqref{eq:identifier} satisfies \begin{align*}
    S_i^k(\delta) = 0, \quad \forall \delta \geq \delta_+ > 0.
\end{align*}As a consequence, the following bound for the threshold at any time $k\geq 0$ holds, \begin{align*}
   \min\left\{\delta_i^0 - \frac{K}{\alpha},-K\left(\frac{1+\alpha}{\alpha}\right)\right\}\leq \delta_i^k\leq \max\left\{\delta_+ +K\left(\frac{1+\alpha}{\alpha}\right), \delta_i^0+\frac{K}{\alpha}\right\}.
\end{align*}
\end{lemma*}
\begin{proof}
First, we prove the upper bound on $\delta_i^k$. There are two distinct cases depending on the initial threshold $\delta_i^0$.
\\

\noindent
\textbf{Case 1:} If $\delta_i^0 < \delta_+$, then there exists and integer $\bar{k}$ such that $\delta^{\bar{k}-1}_i<\delta_+$ and $\delta_i^j \geq \delta_+$ for all {$j \in \J=\{\bar{k}, \dots, k'\}$}.
This implies that 
$S_i^j(\delta_i^j)=0$, for all {$j \in \J$}, hence,
${L_i^{\bar{k}+j'} \leq(1-\alpha)^{j'}}$ for all {$0 \leq j'  \leq k'-\bar{k}$}.
Moreover, due to the fact that $\delta_i^{\bar{k}-1}<\delta_{+}$, this implies that
${\delta_i^{\bar{k}} \leq \delta_{+}+K}$. Consequently, from \eqref{eq:integral_law}, we have
$$ \delta_i^{\bar{k}+1} \leq \delta_+ + K + K$$  and
$$\delta_i^{\bar{k}+2} \leq \delta_+ + K+K+K(1-\alpha), $$ which leads to
$$\delta_i^{\bar{k} + j'} \leq \delta_+ +K+K \sum_{n=0}^{j'-1}(1-\alpha)^{n}.$$ Finally, noting that $\sum_{n = 0}^{j' -1 } (1-\alpha)^n < \sum_{n = 0}^\infty (1-\alpha)^n = \frac{1}{\alpha}$, we have
$$ \delta_i^{\bar{k} + j'} \leq \delta_+ + K\left( \frac{\alpha + 1}{\alpha}\right), \quad {\bar{k} \leq \bar{k}+j'  \leq k'}. $$

\noindent
We apply an induction argument on $k$ and conclude $\delta_i^k \leq \delta_+ + K\left( \frac{\alpha + 1}{\alpha}\right)$, for all {$0\leq k\leq k'$}. {Note that, if $\delta^{k}_i<\delta_+$ as $k>k'$, similar reasoning applies as in the initial procedure $\left(\delta_i^0<\delta_{+}\right)$. Therefore, the upper bound holds for all  $k\geq 0$.}

\noindent
\textbf{Case 2:} If $\delta_i^0 \geq \delta_+$, then $S_i^k(\delta_i^k) = 0$, for all {$k\leq \bar{k}$}, for some $\bar{k}$. This implies that
     $L_i^k \leq (1-\alpha)^{k-1}$
and, therefore
    $$\delta_i^k \leq \delta_i^0 + K + K(1-\alpha) + \cdots + K(1-\alpha)^{k-1} \leq \delta_i^0 + \frac{K}{\alpha},$$
for all $0\leq k\leq \bar{k}$. {Note that, if $\delta^{k}_i<\delta_+$ as $k>\bar{k}$, similar reasoning applies as in the previous case $\left(\delta_i^0<\delta_{+}\right)$.}

Combining the different cases concludes the proof of the following upper bound, for all $k\geq 0$,
\begin{align*}
    \delta_i^k\leq \max\left\{\delta_+ +K\left(\frac{1+\alpha}{\alpha}\right), \delta_i^0+\frac{K}{\alpha}\right\}.
\end{align*}

\noindent
Finally, we prove the lower bound on $\delta_i^k$. We consider the situation where $\delta_i^k < 0$ for some $k$ (otherwise the lower bound is trivially satisfied). This can be investigated via two cases;\\

\noindent
\textbf{Case 1:} 
If $\delta_i^0 \geq 0$,  
then there exists an integer $\bar{k}$ such that $\delta_i^{\bar{k}-1} \geq 0$ and $\delta_i^{j} < 0$ for all $j \in \J = \{ \bar{k},\ldots, k'\}$. Moreover, $|\delta_i^{j+1} - \delta_i^{j}| \leq K$ for all $j \in \J$. This implies that $\delta_i^{j+1} \geq -K$ and $L_i^{j} < \bar{L}_i$ for all $j \in \J$. Also, we have $S_i^j(\delta_i^j)=1$ for all $ j \in \J $, which implies that 
\begin{align*}
    & L_i^{\bar{k}+j'} \geq (1-\alpha)^{j'}L_i^{\bar{k}} + \alpha \sum_{n=0}^{j'-1}(1-\alpha)^n \geq \alpha \sum_{n=0}^{j'-1}(1-\alpha)^n = 1 - (1 - \alpha)^{j'},
\end{align*}
for all $1\leq j'\leq k'-\bar{k}$. Noting that $\delta_i^{\bar{k}+1} \geq -K$, we recall \eqref{eq:integral_law} to get
$$\delta_i^{\bar{k}+j'} \geq -K +K \sum_{n=0}^{j'-1}\left(L_i^{\bar{k}+n} - \bar{L}_i\right).$$
Moreover, since $L_i^{\bar{k}+n} \geq 1 - (1 - \alpha)^{n}$ we have
    $$\delta_i^{\bar{k}+j'} \geq -K + K \sum_{n=0}^{j'-1}\left(1 - (1-\alpha)^n - \bar{L}_i\right).$$
Additionally, $\bar{L}_i \leq 1$ and $\sum_{n = 0}^{j' -1 } (1-\alpha)^n <  \frac{1}{\alpha}$ implies 
    $$\delta_i^{\bar{k}+j'} \geq -K - K \sum_{n=0}^{j'-1}(1-\alpha)^n \geq -K-\frac{K}{\alpha},$$
for all $1\leq j'\leq k'-\bar{k}$. We apply an induction argument on $k$ and conclude $\delta_i^k \geq -K-\frac{K}{\alpha} $ for all {$0 \leq k\leq k'$}. Note that, if $\delta^{j'}_i>0$ as $j'>k'$, similar reasoning applies as in the initial procedure ($\delta_i^0 \geq 0$). Therefore, the lower bound holds for all  $k\geq 0$.
\\
\noindent
\noindent\textbf{Case 2:} If $\delta_i^0 < 0$, and  $\delta_i^{k}<0$  for all $1\leq k\leq \bar{k}$, then following the same procedure as above concludes

$$\delta_i^{k} \geq \delta_i^0 - \frac{K}{\alpha},$$
for all $1\leq k\leq \bar{k}$. {Note that, if $\delta_i^{k}\geq0$ as $k> \bar{k}$, similar reasoning applies as in the previous case. Therefore, the lower bound holds for all  $k\geq 0$.}
\noindent

Combining the different cases concludes the proof of the following lower bound, for all $k\geq 0$,
\begin{align*}
   \min\left\{\delta_i^0 - \frac{K}{\alpha},-K\left(\frac{1+\alpha}{\alpha}\right)\right\}\leq \delta_i^k.
\end{align*}

\end{proof}

\subsection{Proof of Lemma~\ref{lem:communicate}}\label{proof:communicate}
We restate Lemma~\ref{lem:communicate} for the convenience of the reader.

\begin{lemma*}
    Let $K>0$ and $\bar{L}_i >0$, then $\underset{k \to \infty}{\lim \sup}$ $ \:S_i^k(\delta_i^k) = 1$.
\end{lemma*}

\begin{proof} 
    We prove the lemma by contradiction. We therefore assume $$\underset{k \to \infty}{\lim\sup}~ S_i^k(\delta_i^k) \neq 1.$$ From the definition \eqref{eq:identifier}, we know that $S_i^k(\delta_i^k) \in \{0,1\}$, and therefore $${\underset{k \to \infty}{\lim\sup}~ S_i^k(\delta_i^k) = 0}.$$ As a result, there exists an integer $N > 0$ such that $S_i^k(\delta_i^k) = 0$ for all $k > N$, which also implies that $\frac{1}{N} \sum_{j=1}^N S_i^j(\delta_i^j) \to 0$. However, from Thm.~\ref{thm:stable}, we have $\frac{1}{N} \sum_{j=1}^N S_i^j(\delta_i^j) \to \bar{L}_i > 0$, which yields the desired contradiction.
\end{proof} 

\fi
\end{document}